\documentclass[openacc]{rstransa}

\usepackage{cite}
\usepackage{amsmath,amssymb,amsfonts}
\usepackage{algorithmic}
\usepackage{graphicx}
\usepackage{textcomp}
\usepackage{url}            
\usepackage{booktabs}       
\usepackage{nicefrac}       
\usepackage{microtype}      
\usepackage{subfig}
\usepackage{booktabs} 
\usepackage{bm} 
\usepackage{bbm}
\usepackage{hyperref}

\usepackage[numbers]{natbib}

\graphicspath{{figs/}}

\begin{document}

\title{Time Series Forecasting With Deep Learning: A Survey
}

\author{
Bryan Lim$^{1}$ and Stefan Zohren$^{1}$}

\address{$^{1}$Department of Engineering Science, University of Oxford, Oxford, UK}

\subject{Deep learning, time series modelling}


\corres{Bryan Lim\\
\email{blim@robots.ox.ac.uk}}

\begin{abstract}
Numerous deep learning architectures have been developed to accommodate the diversity of time series datasets across different domains. In this article, we survey common encoder and decoder designs used in both one-step-ahead and multi-horizon time series forecasting -- describing how temporal information is incorporated into predictions by each model. Next, we highlight recent developments in hybrid deep learning models, which combine well-studied statistical models with neural network components to improve pure methods in either category. Lastly, we outline some ways in which deep learning can also facilitate decision support with time series data.
\end{abstract}

\begin{fmtext}
\vspace{-1cm}
\section{Introduction}
Time series modelling has historically been a key area of academic research -- forming an integral part of applications in topics such as climate modelling \cite{time_series_climate_ex}, biological sciences \cite{time_series_bio_ex} and medicine \cite{time_series_med_ex}, as well as commercial decision making in retail \cite{RetailDemandExample} and finance \cite{vol_forecasting_finance_ex} to name a few. While traditional methods have focused on parametric models informed by domain expertise -- such as autoregressive (AR) \cite{BoxJenkins}, exponential smoothing \cite{exponential_smoothing,HoltsWinters} or structural time series models \cite{structural_time_series} -- modern machine learning methods provide a means to learn temporal dynamics in a purely data-driven manner \cite{ml_for_time_series}. With the increasing data availability and computing power in recent times, machine learning has become a vital part of the next generation of time series forecasting models.

Deep learning in particular has gained popularity in recent times, inspired by notable achievements in image classification \cite{image_net_ex}, natural language processing \cite{bert_nlp} and reinforcement learning \cite{alpha_go_ex}. By incorporating bespoke architectural assumptions -- or inductive biases \cite{inductive_bias} -- that reflect the nuances of underlying datasets, deep neural networks are able to learn complex data representations \cite{representation_learning}, which alleviates the need for manual feature engineering and model design. The availability of open-source backpropagation frameworks \cite{tensorflow,pytorch} has also simplified the network training, allowing for the customisation for network components and loss functions.
\end{fmtext}

\maketitle
Given the diversity of time-series problems across various domains, numerous neural network design choices have emerged. In this article, we summarise the common approaches to time series prediction using deep neural networks. Firstly, we describe the state-of-the-art techniques available for common forecasting problems -- such as multi-horizon forecasting and uncertainty estimation. Secondly, we analyse the emergence of a new trend in hybrid models, which combine both domain-specific quantitative models with deep learning components to improve forecasting performance. Next, we outline two key approaches in which neural networks can be used to facilitate decision support, specifically through methods in interpretability and counterfactual prediction. Finally, we conclude with some promising future research directions in deep learning for time series prediction -- specifically in the form of continuous-time and hierarchical models. 

While we endeavour to provide a comprehensive overview of modern methods in deep learning, we note that our survey is by no means all-encompassing. Indeed, a rich body of literature exists for automated approaches to time series forecasting - including automatic parametric model selection \cite{forecast_r_package}, and traditional machine learning methods such as kernel regression \cite{kernel_regression} and support vector regression \cite{svm}. In addition, Gaussian processes \cite{gps} have been extensively used for time series prediction -- with recent extensions including deep Gaussian processes \cite{deep_gps}, and parallels in deep learning via neural processes \cite{neural_processes}. Furthermore, older models of neural networks have been used historically in time series applications, as seen in \cite{time_delay_nns} and \cite{old_ts_nn}.

\section{Deep Learning Architectures for Time Series Forecasting}

Time series forecasting models predict future values of a target $y_{i,t}$ for a given entity $i$ at time $t$. Each entity represents a logical grouping of temporal information -- such as measurements from individual weather stations in climatology, or vital signs from different patients in medicine -- and can be observed at the same time. In the simplest case, one-step-ahead forecasting models take the form:
\begin{equation}
    \hat{y}_{i,t+1} = f(y_{i,t-k:t}, \bm{x}_{i,t-k:t}, \bm{s}_i),
    \label{eqn:one_step_pred}
\end{equation}
where $\hat{y}_{i,t+1}$ is the model forecast,  $y_{i,t-k:t} = \{y_{i,t-k}, \dots,  y_{i,t} \}$, $\bm{x}_{i,t-k:t} = \{\bm{x}_{i,t-k},  \dots,  \bm{x}_{i,t} \}$ are observations of the target and exogenous inputs respectively over a look-back window $k$, $s_i$ is static metadata associated with the entity (e.g. sensor location), and $f(.)$ is the prediction function learnt by the model. While we focus on univariate forecasting in this survey (i.e. 1-D targets), we note that the same components can be extended to multivariate models without loss of generality \cite{DeepGLO,deep_quantile_copulas,yaguang_iclr_traffic,deep_forecast,multivariate_ts_copula}. For notational simplicity, we omit the entity index $i$ in subsequent sections unless explicitly required.

\subsection{Basic Building Blocks}
\begin{figure}[bthp]
\vspace{-0.5cm}
\subfloat[CNN Model.\label{fig:cnn}]{\includegraphics[width=0.3\linewidth]{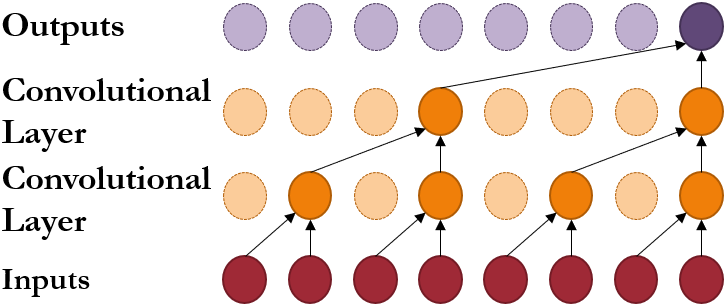}} \hfill
\subfloat[RNN Model.\label{fig:rnn}]{\includegraphics[width=0.3\linewidth]{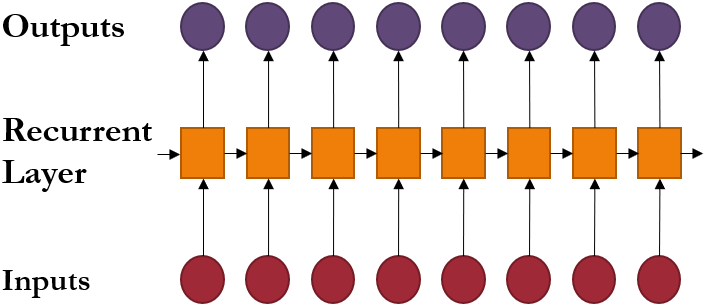}}  \hfill
\subfloat[Attention-based Model.\label{fig:attn}]{\includegraphics[width=0.35\linewidth]{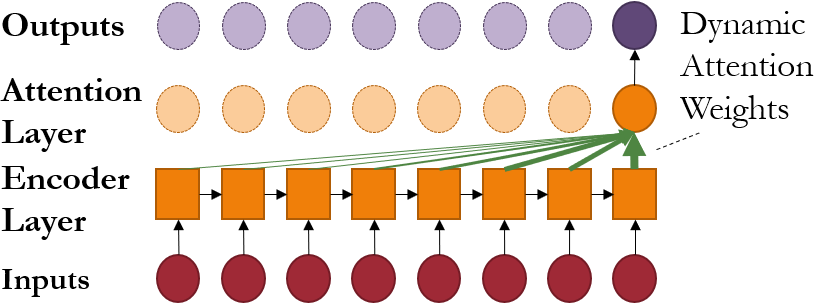}}
\caption{Incorporating temporal information using different encoder architectures. }
\label{fig:model_examples}
\vspace{-0.5cm}
\end{figure}

Deep neural networks learn predictive relationships by using a series of non-linear layers to construct intermediate feature representations \cite{representation_learning}. In time series settings, this can be viewed as encoding relevant historical information into a latent variable $\bm{z}_t$, with the final forecast produced using $\bm{z}_t$ alone:
\begin{align}
    f(y_{t-k:t}, \bm{x}_{t-k:t}, \bm{s}) = g_{\mathrm{dec}}(\bm{z}_t), \label{eqn:decoder}\\
    \bm{z}_t = g_{\mathrm{enc}}(y_{t-k:t}, \bm{x}_{t-k:t}, \bm{s}), 
\end{align}
where $g_{\mathrm{enc}}(.)$, $g_{\mathrm{dec}}(.)$ are encoder and decoder functions respectively, and recalling that that subscript $i$ from Equation \eqref{eqn:one_step_pred} been removed to simplify notation (e.g. $y_{i,t}$ replaced by $y_t$). These encoders and decoders hence form the basic building blocks of deep learning architectures, with the choice of network determining the types of relationships that can be learnt by our model. In this section, we examine modern design choices for encoders, as overviewed in Figure \ref{fig:model_examples}, and their relationship to traditional temporal models. In addition, we explore common network outputs and loss functions used in time series forecasting applications. 

\subsubsection{Convolutional Neural Networks}
Traditionally designed for image datasets, convolutional neural networks (CNNs) extract local relationships that are invariant across spatial dimensions \cite{DeepLearningBook,image_net_ex}. To adapt CNNs to time series datasets, researchers utilise multiple layers of causal convolutions \cite{wavenet,temporal_cnn,cnn_overview} -- i.e. convolutional filters designed to ensure only past information is used for forecasting. For an intermediate feature at hidden layer $l$, each causal convolutional filter takes the form below:
\begin{align}
    \bm{h}_t^{l+1} &= A\bigg(\left(\bm{W} * \bm{h} \right) (l, t)\bigg), \\
    \left(\bm{W} * \bm{h} \right) (l, t) &= \sum_{\tau=0}^{k} \bm{W}(l, \tau)  \bm{h}_{t-\tau}^{l},
    \label{eqn:cnn_conv}
\end{align}
where $\bm{h}_t^{l} \in \mathbb{R}^{\mathcal{H}_{in}}$ is an intermediate state at layer $l$ at time $t$, $*$ is the convolution operator, $\bm{W}(l, \tau) \in \mathbb{R}^{\mathcal{H}_{out}\times \mathcal{H}_{in}}$ is a fixed filter weight at layer $l$, and $A(.)$ is an activation function, such as a sigmoid function, representing any architecture-specific non-linear processing. For CNNs that use a total of L convolutional layers, we note that the encoder output is then $\bm{z}_t=\bm{h}^L_t$.

Considering the 1-D case, we can see that Equation \eqref{eqn:cnn_conv} bears a strong resemblance to finite impulse response (FIR) filters in digital signal processing \cite{dsp}. This leads to two key implications for temporal relationships learnt by CNNs. Firstly, in line with the spatial invariance assumptions for standard CNNs, temporal CNNs assume that relationships are time-invariant -- using the same set of filter weights at each time step and across all time. In addition, CNNs are only able to use inputs within its defined lookback window, or receptive field, to make forecasts. As such, the receptive field size $k$ needs to be tuned carefully to ensure that the model can make use of all relevant historical information. It is worth noting that a single causal CNN layer with a linear activation function is equivalent to an auto-regressive (AR) model.
\paragraph{Dilated Convolutions} Using standard convolutional layers can be computationally challenging where long-term dependencies are significant, as the number of parameters scales directly with the size of the receptive field. To alleviate this, modern architectures frequently make use of dilated covolutional layers \cite{wavenet,temporal_cnn}, which extend Equation \eqref{eqn:cnn_conv} as below:
\begin{align}
    \left(\bm{W} * \bm{h} \right) (l, t, d_l) &= \sum_{\tau=0}^{\lfloor k/d_l \rfloor } \bm{W}(l, \tau)  \bm{h}_{t-d_l\tau }^{l},
    \label{eqn:dilated_convs}
\end{align}
where $\lfloor . \rfloor$ is the floor operator and $d_l$ is a layer-specific dilation rate. Dilated convolutions can hence be interpreted as convolutions of a down-sampled version of the lower layer features -- reducing resolution to incorporate information from the distant past. As such, by increasing the dilation rate with each layer, dilated convolutions can gradually aggregate information at different time blocks, allowing for more history to be used in an efficient manner.  With the WaveNet architecture of \cite{wavenet} for instance,  dilation rates are increased in powers of 2 with adjacent time blocks aggregated in each layer -- allowing for $2^l$ time steps to be used at layer $l$ as shown in Figure \ref{fig:cnn}.

\subsubsection{Recurrent Neural Networks}
Recurrent neural networks (RNNs) have historically been used in sequence modelling \cite{DeepLearningBook}, with strong results on a variety of natural language processing tasks \cite{nlp}. Given the natural interpretation of time series data as sequences of inputs and targets, many RNN-based architectures have been developed for temporal forecasting applications \cite{DeepAR,DSSM,rnf,DeepFactorsForForecasting}. At its core, RNN cells contain an internal memory state which acts as a compact summary of past information. The memory state is recursively updated with new observations at each time step as shown in Figure \ref{fig:rnn}, i.e.:
\begin{align}
    \bm{z}_t &= \nu\left(\bm{z}_{t-1}, y_t, \bm{x}_{t}, \bm{s}\right),
\end{align}
Where $\bm{z}_t \in \mathbb{R}^{\mathcal{H}}$ here is the hidden internal state of the RNN, and $\nu(.)$ is the learnt memory update function. For instance, the Elman RNN \cite{elman_rnn}, one of the simplest RNN variants, would take the form below:
\begin{align}
    y_{t+1} &= \gamma_y(\bm{W}_y \bm{z}_t + \bm{b}_y), \\
    \bm{z}_t &=  \gamma_z(\bm{W}_{z_1} \bm{z}_{t-1} + \bm{W}_{z_2} y_t + \bm{W}_{z_3} \bm{x}_{t} + \bm{W}_{z_4} \bm{s} + \bm{b}_z),
    \label{eqn:rnn_updates}
\end{align}
Where $\bm{W}_., \bm{b}_.$ are the linear weights and biases of the network respectively, and $\gamma_y(.), \gamma_z(.)$ are network activation functions.
Note that RNNs do not require the explicit specification of a lookback window as per the CNN case. From a signal processing perspective, the main recurrent layer -- i.e. Equation \eqref{eqn:rnn_updates} -- thus resembles a non-linear version of infinite impulse response (IIR) filters. 

\paragraph{Long Short-term Memory} Due to the infinite lookback window, older variants of RNNs can suffer from limitations in learning long-range dependencies in the data \cite{bengio_long_term_dependencies,long_term_dep_rnn} -- due to issues with exploding and vanishing gradients \cite{DeepLearningBook}. Intuitively, this can be seen as a form of resonance in the memory state. Long Short-Term Memory networks (LSTMs) \cite{LSTM} were hence developed to address these limitations, by improving gradient flow within the network. This is achieved through the use of a cell state $\bm{c}_t$ which stores long-term information, modulated through a series of gates as below:
\begin{align}
    \text{\textit{Input gate}:}& &\bm{i}_t &= \sigma(\bm{W}_{i_1} \bm{z}_{t-1} + \bm{W}_{i_2} y_t + \bm{W}_{i_3} \bm{x}_{t} + \bm{W}_{i_4} \bm{s} + \bm{b}_i),& \\
    \text{\textit{Output gate}:}& & \bm{o}_t &= \sigma(\bm{W}_{o_1} \bm{z}_{t-1} + \bm{W}_{o_2} y_t + \bm{W}_{o_3} \bm{x}_{t} + \bm{W}_{o_4} \bm{s} + \bm{b}_o), &\\
    \text{\textit{Forget gate}:}& & \bm{f}_t &= \sigma(\bm{W}_{f_1} \bm{z}_{t-1} + \bm{W}_{f_2} y_t + \bm{W}_{f_3} \bm{x}_{t} + \bm{W}_{f_4} \bm{s} + \bm{b}_f),  &
\end{align}
 where $\bm{z}_{t-1}$ is the hidden state of the LSTM, and $\sigma(.)$ is the sigmoid activation function. The gates modify the hidden and cell states of the LSTM as below:
 \begin{align}
    \text{\textit{Hidden state}:} &&\bm{z}_t &= \bm{o}_t \odot \text{tanh}(\bm{c}_t), &\\ 
    \text{\textit{Cell state}:} &&\bm{c}_t &= \bm{f}_t \odot \bm{c}_{t-1}  &\nonumber\\
                                &&& + \bm{i}_t \odot \text{tanh}(\bm{W}_{c_1}\bm{z}_{t-1} + \bm{W}_{c_2} y_t + \bm{W}_{c_3} \bm{x}_{t} + \bm{W}_{c_4} \bm{s} + \bm{b}_c), &
\end{align}
Where $\odot$ is the element-wise (Hadamard) product, and $\text{tanh}(.)$ is the tanh activation function.
\paragraph{Relationship to Bayesian Filtering} As examined in \cite{rnf}, Bayesian filters \cite{bayesian_filtering} and RNNs are both similar in their maintenance of a hidden state which is recursively updated over time. For Bayesian filters, such as the Kalman filter \cite{Kalman_1960}, inference is performed by updating the sufficient statistics of the latent state -- using a series of state transition and error correction steps. As the Bayesian filtering steps use deterministic equations to modify sufficient statistics, the RNN can be viewed as a simultaneous approximation of both steps -- with the memory vector containing all relevant information required for prediction.

\subsubsection{Attention Mechanisms} 
The development of attention mechanisms \cite{attention_1_ex, attention_2_ex} has also lead to improvements in long-term dependency learning -- with Transformer architectures achieving state-of-the-art performance in multiple natural language processing applications \cite{transformer,transformer_xl, bert_nlp}. Attention layers aggregate temporal features using dynamically generated weights (see Figure \ref{fig:attn}), allowing the network to directly focus on significant time steps in the past -- even if they are very far back in the lookback window. Conceptually, attention is a mechanism for a key-value lookup based on a given query \cite{attention_key_query}, taking the form below:
\begin{align}
    \bm{h}_t = \sum_{\tau=0}^k\alpha(\bm{\kappa}_t, \bm{q}_\tau) \bm{v}_{t-\tau},
    \label{eqn:attention}
\end{align}
Where the key $\bm{\kappa}_t$, query $\bm{q}_\tau$ and value $\bm{v}_{t-\tau}$ are intermediate features produced at different time steps by lower levels of the network. Furthermore, $\alpha(\bm{\kappa}_t, \bm{q}_\tau) \in [0, 1]$ is the attention weight for $t-\tau$ generated at time $t$, and $\bm{h}_t$ is the context vector output of the attention layer. Note that multiple attention layers can also be used together as per the CNN case, with the output from the final layer forming the encoded latent variable $\bm{z}_t$.

Recent work has also demonstrated the benefits of using attention mechanisms in time series forecasting applications, with improved performance over comparable recurrent networks \cite{JDLstm,LiTransformer,tft}. For instance, \cite{JDLstm} use attention to aggregate features extracted by RNN encoders, with attention weights produced as below:
\begin{align}
    \bm{\alpha}(t) &= \text{softmax}(\bm{\eta}_t), \\
    \bm{\eta}_t &= \mathbf{W}_{\eta_1} \text{tanh}(\mathbf{W}_{\eta_2} \bm{\kappa}_{t-1} + \mathbf{W}_{\eta_3} \bm{q}_\tau  + \bm{b}_\eta),
\end{align}
where $\bm{\alpha}(t) = [\alpha(t,0), \dots \alpha(t,k)]$ is a vector of attention weights, $\bm{\kappa}_{t-1}, \bm{q}_t$ are outputs from LSTM encoders used for feature extraction, and $\text{softmax}(.)$ is the softmax activation function. More recently, Transformer architectures have also been considered in \cite{LiTransformer,tft}, which apply scalar-dot product self-attention \cite{transformer} to features extracted within the lookback window. From a time series modelling perspective, attention provides two key benefits. Firstly, networks with attention are able to directly attend to any significant events that occur. In retail forecasting applications, for example, this includes holiday or promotional periods which can have a positive effect on sales. Secondly, as shown in \cite{tft}, attention-based networks can also learn regime-specific temporal dynamics -- by using distinct attention weight patterns for each regime.

\subsubsection{Outputs and Loss Functions}
\label{sec:outputs_and_loss_fxns}
Given the flexibility of neural networks, deep neural networks have been used to model both discrete \cite{retain} and continuous \cite{DeepAR,MQRNN} targets -- by customising of decoder and output layer of the neural network to match the desired target type. In one-step-ahead prediction problems, this can be as simple as combining a linear transformation of encoder outputs (i.e. Equation \eqref{eqn:decoder}) together with an appropriate output activation for the target. Regardless of the form of the target, predictions can be further divided into two different categories -- point estimates and probabilistic forecasts.

\paragraph{Point Estimates} A common approach to forecasting is to determine the expected value of a future target. This essentially involves reformulating the problem to a classification task for discrete outputs (e.g. forecasting future events), and regression task for continuous outputs -- using the encoders described above. For the binary classification case, the final layer of the decoder then features a linear layer with a sigmoid activation function -- allowing the network to predict the probability of event occurrence at a given time step. For one-step-ahead forecasts of binary and continuous targets, networks are trained using binary cross-entropy and mean square error loss functions respectively:
\begin{align} 
    \mathcal{L}_{classification} &= -\frac{1}{T} \sum_{t=1}^T  y_t \log(\hat{y}_t) + (1-y_t) \log(1-\hat{y}_t) \\
    \mathcal{L}_{regression} &= \frac{1}{T} \sum_{t=1}^T \left( y_t - \hat{y}_t \right)^2
\end{align}

While the loss functions above are the most common across applications, we note that the flexibility of neural networks also allows for more complex losses to be adopted - e.g. losses for quantile regression \cite{MQRNN} and multinomial classification \cite{wavenet}.

\paragraph{Probabilistic Outputs} While point estimates are crucial to predicting the future value of a target, understanding the uncertainty of a model's forecast can be useful for decision makers in different domains. When forecast uncertainties are wide, for instance, model users can exercise more caution when incorporating predictions into their decision making, or alternatively rely on other sources of information. In some applications, such as financial risk management, having access to the full predictive distribution will allow decision makers to optimise their actions in the presence of rare events -- e.g. allowing risk managers to insulate portfolios against market crashes. 

A common way to model uncertainties is to use deep neural networks to generate parameters of known distributions \cite{DeepAR,deep_quantile_copulas,DSSM}. For example, Gaussian distributions are typically used for forecasting problems with continuous targets, with the networks outputting means and variance parameters for the predictive distributions at each step as below:
\begin{align}
    y_{t+\tau} &\sim N(\mu(t,\tau), \zeta(t, \tau)^2), \label{eqn:gaussian_sim} \\
    \mu(t,\tau) &= \bm{W}_\mu \bm{h}^L_t + \bm{b}_\mu, \\
    \zeta(t, \tau) &= \text{softplus}(\bm{W}_\Sigma\bm{h}^L_t + \bm {b}_\Sigma ),
\end{align}
where $\bm{h}^L_t$ is the final layer of the network, and $\text{softplus}(.)$ is the softplus activation function to ensure that standard deviations take only positive values.

\subsection{Multi-horizon Forecasting Models}
\begin{figure}[bthp]
\subfloat[Iterative Methods \label{fig:iterative}]{\includegraphics[width=0.475\linewidth]{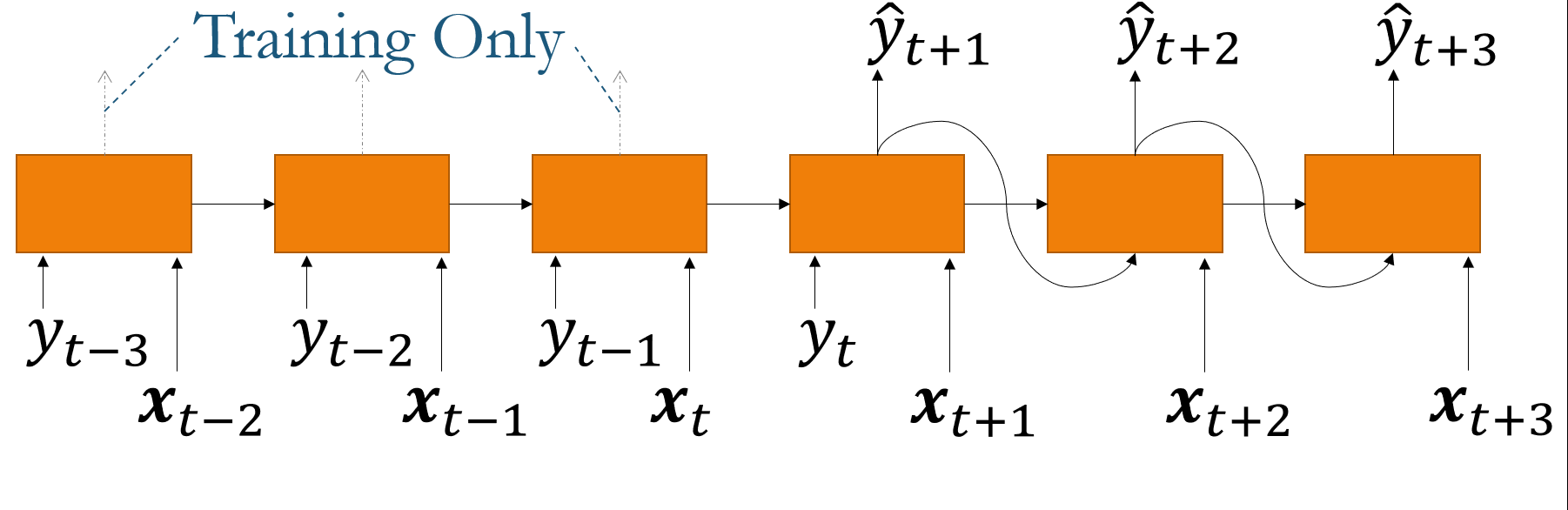}} \hfill
\subfloat[Direct Methods \label{fig:direct}]{\includegraphics[width=0.475\linewidth]{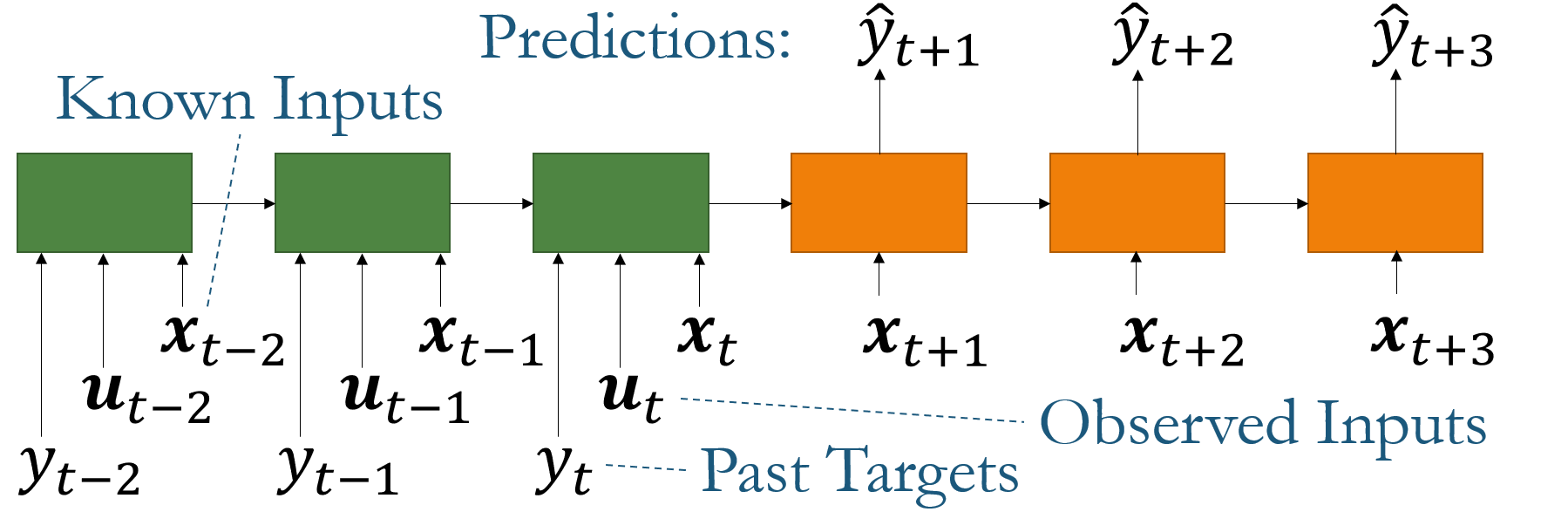}}
\caption{Main types of multi-horizon forecasting models. Colours used to distinguish between model weights -- with iterative models using a common model across the entire horizon and direct methods taking a sequence-to-sequence approach.}
\label{fig:multihorizon}
\vspace{-0.75cm}
\end{figure}

In many applications, it is often beneficial to have access to predictive estimates at multiple points in the future -- allowing decision makers to visualise trends over a future horizon, and optimise their actions across the entire path. From a statistical perspective, multi-horizon forecasting can be viewed as a slight modification of one-step-ahead prediction problem (i.e. Equation \eqref{eqn:one_step_pred}) as below:
\begin{equation}
    \hat{y}_{t+\tau} = f(y_{t-k:t}, \bm{x}_{t-k:t}, \bm{u}_{t-k:t+\tau}, \bm{s}, \tau),
    \label{eqn:multistep_pred}
\end{equation}
where $\tau \in \{1, \dots, \tau_{max}\}$ is a discrete forecast horizon, $\bm{u}_t$ are known future inputs (e.g. date information, such as the day-of-week or month) across the entire horizon, and $\bm{x}_t$ are inputs that can only be observed historically.  In line with traditional econometric approaches \cite{MultistepMethods,DirectVsIterated}, deep learning architectures for multi-horizon forecasting can be divided into iterative and direct methods -- as shown in Figure \ref{fig:multihorizon} and described in detail below. 
\subsubsection{Iterative Methods}
Iterative approaches to multi-horizon forecasting typically make use of autoregressive deep learning architectures \cite{DeepAR,rnf,LiTransformer,DeepFactorsForForecasting} -- producing multi-horizon forecasts by recursively feeding samples of the target into future time steps (see Figure \ref{fig:iterative}). By repeating the procedure to generate multiple trajectories, forecasts are then produced using the sampling distributions for target values at each step. For instance, predictive means can be obtained using the Monte Carlo estimate $\hat{y}_{t+\tau} = \sum_{j=1}^J \tilde{y}_{t+\tau}^{(j)} / J$, where $\tilde{y}_{t+\tau}^{(j)}$ is a sample taken based on the model of Equation \eqref{eqn:gaussian_sim}. As autoregressive models are trained in the exact same fashion as one-step-ahead prediction models (i.e. via backpropagation through time), the iterative approach allows for the easy generalisation of standard models to multi-step forecasting. However, as a small amount of error is produced at each time step, the recursive structure of iterative methods can potentially lead to large error accumulations over longer forecasting horizons. In addition, iterative methods assume that all inputs but the target are known at run-time -- requiring only samples of the target to be fed into future time steps. This can be a limitation in many practical scenarios where observed inputs exist, motivating the need for more flexible methods.

\subsubsection{Direct Methods}
Direct methods alleviate the issues with iterative methods by producing forecasts directly using all available inputs. They typically make use of sequence-to-sequence architectures \cite{MQRNN,JDLstm,tft}, using an encoder to summarise past information (i.e. targets, observed inputs and a priori known inputs), and a decoder to combine them with known future inputs -- as depicted in Figure \ref{fig:direct}. As described in \cite{M3Analysis}, alternative approach is to use simpler models to directly produce a fixed-length vector matching the desired forecast horizon. This, however, does require the specification of a maximum forecast horizon (i.e. $\tau_{max}$), with predictions made only at the predefined discrete intervals.


\if false
\subsection{Methods for Probabilistic Modelling}
\subsubsection{Modifying Network Outputs}
\paragraph{Parameterised Distributions}
\paragraph{Quantile Forecasts}
\paragraph{Ordinal Regression}
\subsubsection{Deep Generative Models}
\subsubsection{Bayesian Neural Networks}
\fi

\section{Incorporating Domain Knowledge with Hybrid Models}
Despite its popularity, the efficacy of machine learning for time series prediction has historically been questioned -- as evidenced by forecasting competitions such as the M-competitions \cite{comp_history}. Prior to the M4 competition of 2018 \cite{M4Competition}, the prevailing wisdom was that sophisticated methods do not produce more accurate forecasts, and simple models with ensembling had a tendency to do better \cite{M1Results,M3Results,M3Analysis}. Two key reasons have been identified to explain the underperformance of machine learning methods. Firstly, the flexibility of machine learning methods can be a double-edged sword -- making them prone to overfitting \cite{M3Analysis}. Hence, simpler models may potentially do better in low data regimes, which are particularly common in forecasting problems with a small number of historical observations (e.g. quarterly macroeconomic forecasts). Secondly, similar to stationarity requirements of statistical models, machine learning models can be sensitive to how inputs are pre-processed \cite{DeepAR,DeepGLO,M3Analysis}, which ensure that data distributions at training and test time are similar.

A recent trend in deep learning has been in developing hybrid models which address these limitations, demonstrating improved performance over pure statistical or machine learning models in a variety of applications \cite{esrnn,DSSM,dmns,hybrid_weather_climate_ex}. Hybrid methods combine well-studied quantitative time series models together with deep learning -- using deep neural networks to generate model parameters at each time step. On the one hand, hybrid models allow domain experts to inform neural network training using prior information -- reducing the hypothesis space of the network and improving generalisation. This is especially useful for small datasets \cite{DSSM}, where there is a greater risk of overfitting for deep learning models. Furthermore, hybrid models allow for the separation of stationary and non-stationary components, and avoid the need for custom input pre-processing. An example of this is the Exponential Smoothing RNN (ES-RNN) \cite{esrnn}, winner of the M4 competition, which uses exponential smoothing to capture non-stationary trends and learns additional effects with the RNN. In general, hybrid models utilise deep neural networks in two manners: a) to encode time-varying parameters for non-probabilistic parametric models  \cite{esrnn,dmns,ar_cnns}, and b) to produce parameters of distributions used by probabilistic models \cite{DSSM,DeepFactorsForForecasting,hybrid_weather_climate_ex}.

\subsection{Non-probabilistic Hybrid Models}
With parametric time series models, forecasting equations are typically defined analytically and provide point forecasts for future targets. Non-probabilistic hybrid models hence modify these forecasting equations to combine statistical and deep learning components. The ES-RNN for example, utilises the update equations of the Holt-Winters exponential smoothing model \cite{HoltsWinters} -- combining multiplicative level and seasonality components with deep learning outputs as below:
\begin{align}
    \hat{y}_{i,t+\tau} &= \exp(\bm{W}_{ES} \bm{h}^{L}_{i,t+\tau} + \bm{b}_{ES}) \times l_{i,t} \times \gamma_{i,t+\tau}, \\
    l_{i,t} &= \beta_{1}^{(i)} y_{i,t}/\gamma_{i,t} + (1-\beta_{1}^{(i)}) l_{i,t-1}, \\
    \gamma_{i,t} &= \beta_{2}^{(i)} y_{i,t}/l_{i,t} + (1-\beta_{2}^{(i)})\gamma_{i,t-\kappa},
\end{align}
where $\bm{h}^L_{i,t+\tau}$ is the final layer of the network for the $\tau$th-step-ahead forecast, $l_{i,t}$ is a level component, $\gamma_{i,t}$ is a seasonality component with period $\kappa$, and $\beta_1^{(i)}, \beta_2^{(i)}$ are entity-specific static coefficients. From the above equations, we can see that the exponential smoothing components ($l_{i,t},\gamma_{i,t}$) handle the broader (e.g. exponential) trends within the datasets, reducing the need for additional input scaling.

\subsection{Probabilistic Hybrid Models} 
Probabilistic hybrid models can also be used in applications where distribution modelling is important -- utilising probabilistic generative models for temporal dynamics such as Gaussian processes \cite{DeepFactorsForForecasting} and linear state space models \cite{DSSM}. Rather than modifying forecasting equations, probabilistic hybrid models use neural networks to produce parameters for predictive distributions at each step. For instance, Deep State Space Models \cite{DSSM} encode time-varying parameters for linear state space models as below -- performing inference via the Kalman filtering equations \cite{Kalman_1960}:
\begin{align}
    y_t &= \bm{a}(\bm{h}^{L}_{i,t+\tau})^T \bm{l}_t + \phi(\bm{h}^{L}_{i,t+\tau}) \epsilon_t, \\
    \bm{l}_t &= \bm{F}(\bm{h}^{L}_{i,t+\tau}) \bm{l}_{t-1} + \bm{q}(\bm{h}^{L}_{i,t+\tau}) + \bm{\Sigma}(\bm{h}^{L}_{i,t+\tau}) \odot \bm{\Sigma}_t, 
\end{align}
where $\bm{l}_t$ is the hidden latent state, $\bm{a}(.)$, $\bm{F}(.)$, $\bm{q}(.)$ are linear transformations of $\bm{h}^{L}_{i,t+\tau}$, $\phi(.)$, $\bm{\Sigma}(.)$ are linear transformations with softmax activations, $\epsilon_t \sim N(0,1)$ is a univariate residual and  $\bm{\Sigma}_t \sim N(0,\mathbb{I})$ is a multivariate normal random variable.

\section{Facilitating Decision Support Using Deep Neural Networks}
Although model builders are mainly concerned with the accuracy of their forecasts, end-users typically use predictions to \textit{guide their future actions}. For instance, doctors can make use of clinical forecasts (e.g. probabilities of disease onset and mortality) to help them prioritise tests to order, formulate a diagnosis and determine a course of treatment. As such, while time series forecasting is a crucial preliminary step, a better understanding of both temporal dynamics and the motivations behind a model's forecast can help users further optimise their actions. In this section, we explore two directions in which neural networks have been extended to facilitate decision support with time series data -- focusing on methods in interpretability and causal inference.

\subsection{Interpretability With Time Series Data}
With the deployment of neural networks in mission-critical applications \cite{causal_interp_2020}, there is a increasing need to understand both \textit{how} and \textit{why} a model makes a certain prediction. Moreover, end-users can have little prior knowledge with regards to the relationships present in their data, with datasets growing in size and complexity in recent times. Given the black-box nature of standard neural network architectures, a new body of research has emerged in methods for interpreting deep learning models. We present a summary below -- referring the reader to dedicated surveys for more in-depth analyses \cite{dl_interp_survey, nature_interp}.

\paragraph{Techniques for Post-hoc Interpretability} Post-hoc interpretable models are developed to interpret trained networks, and helping to identify important features or examples without modifying the original weights.  Methods can mainly be divided into two main categories. Firstly, one possible approach is to apply simpler interpretable surrogate models between the inputs and outputs of the neural network, and rely on the approximate model to provide explanations. For instance, Local Interpretable Model-Agnostic Explanations (LIME) \cite{LIME} identify relevant features by fitting instance-specific linear models to perturbations of the input, with the linear coefficients providing a measure of importance. Shapley additive explanations (SHAP) \cite{SHAP} provide another surrogate approach, which utilises Shapley values from cooperative game theory to identify important features across the dataset. Next, gradient-based method -- such as saliency maps \cite{saliency_maps, saliency_ts} and influence functions \cite{influence_functions} -- have been proposed, which analyse network gradients to determine which input features have the greatest impact on loss functions. While post-hoc interpretability methods can help with feature attributions, they typically ignore any sequential dependencies between inputs -- making it difficult to apply them to complex time series datasets.

\paragraph{Inherent Interpretability with Attention Weights} An alternative approach is to directly design architectures with explainable components, typically in the form of strategically placed attention layers.  As attention weights are produced as outputs from a softmax layer, the weights are constrained to sum to 1, i.e. $\sum_{\tau=0}^k\alpha(t, \tau) = 1$. For time series models, the outputs of Equation \eqref{eqn:attention} can hence also be interpreted as a weighted average over temporal features, using the weights supplied by the attention layer at each step. An analysis of attention weights can then be used to understand the relative importance of features at each time step. Instance-wise interpretability studies have been performed in \cite{kdd_medical_attn,retain,LiTransformer}, where the authors used specific examples to show how the magnitudes of  $\alpha(t, \tau)$ can indicate which time points were most significant for predictions. By analysing distributions of attention vectors across time, \cite{tft} also shows how attention mechanisms can be used to identify persistent temporal relationships -- such as seasonal patterns -- in the dataset.

\subsection{Counterfactual Predictions \& Causal Inference Over Time}
In addition to understanding the relationships learnt by the networks, deep learning can also help to facilitate decision support by producing predictions outside of their observational datasets, or counterfactual forecasts. Counterfactual predictions are particularly useful for scenario analysis applications -- allowing users to evaluate how different sets of actions can impact target trajectories. This can be useful both from a historical angle, i.e. determining what would have happened if a different set of circumstances had occurred, and from a forecasting perspective, i.e. determining which actions to take to optimise future outcomes. 

While a large class of deep learning methods exists for estimating causal effects in static settings \cite{GANITE, DeepIV,CounterfactualNetworksPropensityDropout}, the key challenge in time series datasets is the presence of time-dependent confounding effects. This arises due to circular dependencies when actions that can affect the target are also conditional on observations of the target. Without any adjusting for time-dependent confounders, straightforward estimations techniques can results in biased results, as shown in \cite{TimeVaryingConfoundingAdjustments}. 
Recently, several methods have emerged to train deep neural networks while adjusting for time-dependent confounding, based on extensions of statistical techniques and the design of new loss functions. With statistical methods, \cite{RMSN} extends the inverse-probability-of-treatment-weighting (IPTW) approach of marginal structural models in epidemiology -- using one set of networks to estimate treatment application probabilities, and a sequence-to-sequence model to learn unbiased predictions. Another approach in \cite{g_net} extends the G-computation framework, jointly modelling distributions of the target and actions using deep learning. In addition, new loss functions have been proposed in \cite{adversarially_balanced_reps}, which adopts domain adversarial training to learn balanced representations of patient history.

\section{Conclusions and Future Directions}
With the growth in data availability and computing power in recent times, deep neural networks architectures have achieved much success in forecasting problems across multiple domains. In this article, we survey the main architectures used for time series forecasting -- highlighting the key building blocks used in neural network design. We examine how they incorporate temporal information for one-step-ahead predictions, and describe how they can be extended for use in multi-horizon forecasting. Furthermore, we outline the recent trend of hybrid deep learning models, which combine statistical and deep learning components to outperform pure methods in either category. Finally, we summarise two ways in which deep learning can be extended to improve decision support over time, focusing on methods in interpretability and counterfactual prediction.

Although a large number of deep learning models have been developed for time series forecasting, some limitations still exist. Firstly, deep neural networks typically require time series to be discretised at regular intervals, making it difficult to forecast datasets where observations can be missing or arrive at random intervals. While some preliminary research on continuous-time models has been done via Neural Ordinary Differential Equations \cite{NeuralODEs}, additional work needs to be done to extend this work for datasets with complex inputs (e.g. static variables) and to benchmark them against existing models. In addition, as mentioned in \cite{M4Example}, time series often have a hierarchical structure with logical groupings between trajectories -- e.g. in retail forecasting, where product sales in the same geography can be affected by common trends. As such, the development of architectures which explicit account for such hierarchies could be an interesting research direction, and potentially improve forecasting performance over existing univariate or multivariate models.\\

\competing{The author(s) declare that they have no competing interests.}
\if false
\ethics{Insert ethics statement here if applicable.}
\dataccess{Insert details of how to access any supporting data here.}
\aucontribute{For manuscripts with two or more authors, insert details of the authors’ contributions here. This should take the form: 'AB carried out the experiments. CD performed the data analysis. EF conceived of and designed the study, and drafted the manuscript All authors read and approved the manuscript'.}
\competing{Insert any competing interests here. If you have no competing interests please state 'The author(s) declare that they have no competing interests’.}
\funding{Insert funding text here.}
\ack{Insert acknowledgment text here.}
\disclaimer{Insert disclaimer text here if applicable.}
\fi
\footnotesize{
\bibliography{bib_general,bib_forecasting,bib_interp,bib_probs,bib_examples}
\bibliographystyle{vancouver}}

\end{document}